\documentclass[conference]{IEEEtran}
\IEEEoverridecommandlockouts
\usepackage{cite}
\usepackage{amsmath,amssymb,amsfonts}
\usepackage{algorithmic}
\usepackage{graphicx}
\usepackage{textcomp}
\usepackage{xcolor}
\usepackage{tabularray}
\usepackage{hyperref}
\def\BibTeX{{\rm B\kern-.05em{\sc i\kern-.025em b}\kern-.08em
    T\kern-.1667em\lower.7ex\hbox{E}\kern-.125emX}}

\usepackage{url}

\begin{document}

\title{
Arabic Handwritten Document OCR Solution with\\ Binarization and Adaptive Scale Fusion Detection
}

\author{
\IEEEauthorblockN{\textbf{Alhossien Waly\textsuperscript{*1} Bassant Tarek\textsuperscript{*2} Ali Feteha\textsuperscript{*3}, Rewan Yehia\textsuperscript{*4} Gasser Amr\textsuperscript{*5} Ahmed Fares\textsuperscript{*\#6}}}
    \IEEEauthorblockA{
        \textsuperscript{\textbf{*}}Computer Science and Engineering Departement, Faculty of Engineering, \\
        Egypt-Japan University of Science \& Technology E-JUST,
        Alexandria 21934, Egypt\\
        email:\{alhossien.waly, bassant.tarek, ali.ibrahim,
        rewan.abubakr, gasser.amr, ahmed.fares\}@ejust.edu.eg\\
        \textsuperscript{\textbf{\#}}Electrical Engineering Department, Faculty of Engineering,\\ Benha University,
        Cairo 11629, Egypt\\
    }
}

\maketitle

\begin{abstract}
The problem of converting images of text into plain text is a widely researched topic in both academia and industry. Arabic handwritten Text Recognation (AHTR) poses additional challenges due to diverse handwriting styles and limited labeled data. In this paper we present a complete OCR pipeline that starts with line segmentation using Differentiable Binarization and Adaptive Scale Fusion techniques to ensure accurate detection of text lines. Following segmentation, a CNN-BiLSTM-CTC architecture is applied to recognize characters.  Our system, trained on the Arabic Multi-Fonts Dataset (AMFDS), achieves a Character Recognition Rate (CRR) of 99.20\% and a Word Recognition Rate (WRR) of 93.75\% on single-word samples containing 7 to 10 characters, along with a CRR of 83.76\% for sentences. These results demonstrate the system’s strong performance in handling Arabic scripts, establishing a new benchmark for AHTR systems.


\end{abstract}

\section{INTRODUCTION}
 Digitizing Handwritten text has huge importance and quite a history in almost all industries With commercial transactions of money, exchange of raw materials, or digitalizing physical products. Arabic Handwritten Text Recognition (AHTR) is very challenging due to the cursive nature of the script, complex ligatures, and variability in handwriting styles. Such problems are, however, more profound in the instance where datasets are not properly labeled; hence, the development of robust Optical Character Recognition (OCR) for handwritten Arabic is indeed a very hard task [\ref{12}], [\ref{13}].

While earlier, early OCR systems did display moderate success with the help of techniques such as Hidden Markov Models (HMM) and Support Vector Machines (SVM) applied on it [\ref{12}], [\ref{13}], they were not able to generalize the rules across distinct styles of writing. Recent progress in deep learning the work of Graves et al. [\ref{4}] presents the CNN-BLSTM-CTC model now much more increases state-of-the-art accuracy for sequence modeling and recognition in complex scripts, such as Arabic.

Effective segmentation is essential for improving handwritten OCR, especially when dealing with the intricate and connected characters of Arabic script. Traditional segmentation methods often face challenges due to the script's overlapping and cursive nature, which results in less-than-reliable recognition accuracy. However, recent innovations, including Differentiable Binarization and Adaptive Scale Fusion implemented in models like DBNet++ have shown substantial improvements. These techniques enhance the resilience of text detection systems by managing variable text scales and preserving critical contextual details, leading to more accurate and efficient processing of handwritten Arabic text Liao et al. [\ref{11}].

This paper introduces a method that merges Differentiable Binarization and Adaptive Scale Fusion to enhance the accuracy of text segmentation, followed by the use of a CNN-BLSTM-CTC OCR engine to ensure reliable recognition. This combined approach effectively tackles the unique challenges posed by Arabic handwritten OCR, offering a flexible and scalable solution for document digitization and data extraction.

\section{RELATED WORK}

The text detection problem has quite a history of contribution either by computer-vision algorithms or deep-learning techniques. Initially, scene text detection techniques would often identify individual characters or components and then combine them into words. Neumann and Matas [\ref{9}] suggested using Extremal Regions (ERs) classification to find characters. An effective sequential selection from the set of Extremal Regions is how they presented the character detection problem. The identified ERs were then categorized into words. 

The field of scene text detection has recently been dominated by deep learning. Depending on how precise the projected target is, the deep learning-based scene text recognition techniques fall into one of three broad categories: segmentation-based, part-based, and regression-based techniques

Following the recently proposed DBNet++ [\ref{11}] which performed more accurately and more efficiently owing to the simple and efficient differentiable binarization algorithm with adaptive fusion outperforming other recent methods. Therefore we choose to build our line segmentation on it.

Handwritten Arabic Optical Character Recognition (OCR) faces unique challenges due to connected characters, overlapping ligatures, and regional variations in writing styles. Over recent years, active research has been conducted in developing robust approaches to Arabic handwritten OCR, with significant progress in deep learning methods.

Early OCR systems relied heavily on handcrafted features and traditional machine learning classifiers. Notable among these was the work of Al-Hajj et al. [\ref{12}], who combined structural and statistical features with Hidden Markov Models (HMMs) to achieve a character accuracy of 92.1\% on a proprietary dataset. Similarly, Elzobi et al. [\ref{13}] utilized Gabor filter-based feature extraction coupled with Support Vector Machine classification, obtaining a 94.3\% character recognition performance on the IFN/ENIT dataset [\ref{14}].

The advent of deep learning marked a paradigm shift in OCR, moving from manual Feature Engineering to end-to-end learning for both feature generation and recognition. A significant breakthrough came with the work of Graves et al. [\ref{4}], who introduced the CNN-BLSTM-CTC architecture. This model combined CNNs for Feature Extraction, Bidirectional Long Short-Term Memory (BLSTM) for Sequence Modeling, and Connectionist Temporal Classification (CTC) for alignment.

Ahmad et al. [\ref{17}] made a notable advancement by applying a CNN-BLSTM-CTC model to the KHATT dataset, achieving a Character Recognition Rate (CRR) of 80.02\%. Other work trained on a custom dataset of over two million word samples from 18 different fonts. When tested on unseen data, the model demonstrated a CRR of 85.15\% on one Word, underscoring its robustness and generalization capability [\ref{28}].

Recent advancements in deep learning have led to the development of Transformer-based models. Yousef et al. [\ref{21}] designed a deep Transformer-based encoder-decoder architecture for Arabic handwritten OCR, outperforming CNN-BLSTM-CTC based systems on several benchmarks, including the KHATT dataset, where it achieved a CER of 3.1\%, representing a 35.4\% relative improvement.

However, the deployment of Transformer in resource-constrained environments remains challenging [\ref{22}]. For this reason, the CNN-BLSTM-CTC architecture continues to be a popular choice for practical applications, offering a balance between recognition accuracy and computational efficiency.

This work presents an Arabic handwritten OCR system based on Differentiable Binarization and Adaptive Scale Fusion text detection and CNN-BLSTM-CTC architecture, enhanced with end-to-end Beam Search. This approach aims to leverage the power of deep learning and language modeling to accurately and effectively recognize characters for real-world applications in document digitization and information retrieval.

\section{Methodology}
\begin{figure*}[ht]
    \centering
    \includegraphics[width=\linewidth]{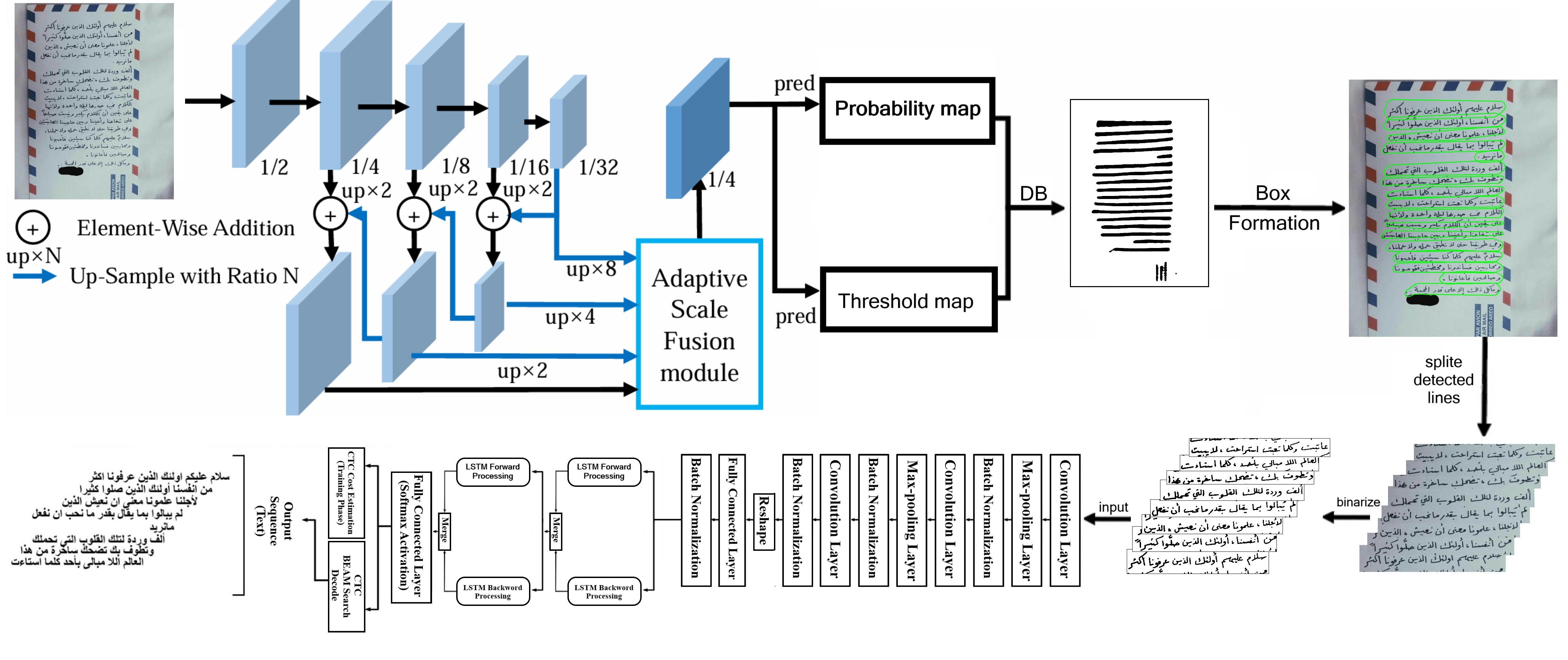}
    \caption{The figure shows the pipeline of the proposed solution starting with DBnet++[\ref{11}] architecture for Line Segmentation of the Arabic Handwritten text, then Binarizing each detected line. The final step is feeding the Binarized lines into the CNN network followed by Bidirectional-LSTM and decoding the output via Connectionist Temporal Classification "CTC"}
    \label{fig: Process illustration}
    \vspace{-10pt}
\end{figure*}

The methodology for this study is methodically outlined in Figure \ref{fig: Process illustration}, which depicts a streamlined pipeline that initiates with line segmentation using the DBNet++ architecture, advances through the Binarization phase, and concludes with Character Recognition via a specially designed CNN-BiLSTM-CTC model. This sequential approach ensures that each phase is meticulously optimized to seamlessly transition to the next, thereby enhancing the accuracy and operational efficiency of our OCR system tailored for Arabic Handwritten texts.
\vspace{-2pt}
\subsection{Line Segmentation}
Our work Method initially Follows the recently proposed DBNet++ [\ref{11}] which performed well on several printed text datasets. As our detection objective "Handwritten Arabic Text" is similar to the printed text, we fine-tuned Universal best-weights trained on ICDAR 2015[\ref{25}], Total-Text [\ref{26}], MSRA-TD500 [\ref{27}], and Chinese Baidu[\ref{29}] using Handwritten text images to add value to the universal weights.
\subsubsection{Model background (DBNet++) [\ref{11}]}
The model starts by feeding the image into the ResNet50 Backbone for Feature Extraction then scales them up until to reach same scale for all to pass the features to The Adaptive Scale Fusion (ASF) module. The ASF generates contextual features to predict probability map and threshold map as in Figure \ref{fig: Process illustration}. 
After that comes the calculation of Approximate Binary maps using a Probability map and a Feature map. During the training period, the supervision is applied on the Probability map, the Threshold
map, and the Approximate Binary map, where the Probability map and the Approximate Binary map share the same Supervision. 
On Testing, Bounding Boxes are derived from either the approximate Binary map or the Probability map via a box formation process.
\subsubsection{Segmentation Dataset}
We used Arabic Documents OCR Dataset [\ref{23}] for our Line Segmentation Task.
The dataset contains 10K printed and handwritten text images split into 12 classes. We only used two classes of the dataset "Handwritten text and official documents" Which suit our problem 1.6K images.
\vspace{-2pt}
\subsection{OCR Engine}
Our Optical Character Recognition (OCR) system is tailored to handle segmented lines of Arabic text. It employs a robust Pipeline that leverages advanced Deep-Learning models to recognize and decode Handwritten Arabic scripts efficiently.

\begin{flushright}
\begin{table}[ht]
\vspace{-10pt}
\caption{Detailed architecture of the OCR Model}
\hspace*{-0.2cm}
\label{tab:model-architecture}
\centering
\begin{tabular}{|c|c|c|}
\hline
\textbf{Layer} & \textbf{Config} & \textbf{Notes} \\
\hline
Convolutional layer& $3 \times 3$, 32 & Activation: ReLU\\
\hline
MaxPooling & $2 \times 2$ & Pooling window \\
\hline
BatchNorm & - & - \\
\hline
\hline
Convolutional layer& $3 \times 3$, 64 & Activation: ReLU\\
\hline
MaxPooling& $2 \times 2$ & Pooling window \\
\hline
BatchNorm & - & - \\
\hline
\hline
Convolutional layer& $3 \times 3$, 128 & Activation: ReLU\\
\hline
BatchNorm & - & - \\
\hline
\hline
Dense & 64 units & Activation: ReLU \\
\hline
BatchNorm & - & - \\
\hline
\hline
Bi-LSTM & 128 units & Return sequences \\
\hline
Bi-LSTM & 256 units & Return sequences \\
\hline
\hline
Dense & Softmax & Classes: \# vocabulary + [blank] \\
\hline
\hline
CTC & - & Connectionist Temporal Classification \\
\hline
\end{tabular}
\end{table}
\end{flushright}

\subsubsection{Model Architecture}
Our OCR engine's architecture is meticulously crafted to optimize the recognition of Arabic handwritten text. It begins with a Convolutional Neural Network (CNN) that processes input images to extract relevant Features. This is followed by a Bidirectional Long Short-Term Memory (Bi-LSTM) network, which excels in capturing the Contextual Dependencies necessary for decoding sequential data. The model employs a Connectionist Temporal Classification (CTC) layer, aligning input sequences with their corresponding labels without requiring explicit Segmentation. As detailed in Table \ref{tab:model-architecture}, this configuration ensures a robust framework capable of handling the complexities of Arabic script recognition.
The selection of model design was determined through experiments on different Hyperparameter and Layers.

\subsubsection{OCR Dataset Preparation}
\begin{figure}[htbp]
    \centering
    \begin{minipage}[b]{0.2\linewidth}
        \centering
        \frame{\includegraphics[height=0.5cm]{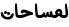}}
    \end{minipage}
    \hspace{0.01\linewidth}
    \begin{minipage}[b]{0.2\linewidth}
        \centering
        \frame{\includegraphics[height=0.5cm]{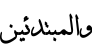}}
    \end{minipage}
    \hspace{0.01\linewidth}
    \begin{minipage}[b]{0.2\linewidth}
        \centering
        \frame{\includegraphics[height=0.4cm]{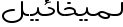}}
    \end{minipage}
    \hspace{0.01\linewidth}
    \begin{minipage}[b]{0.2\linewidth}
        \centering
        \frame{\includegraphics[height=0.5cm]{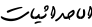}}
    \end{minipage}
    \hspace{0.01\linewidth}
    \begin{minipage}[b]{0.2\linewidth}
        \centering
        \frame{\includegraphics[height=0.5cm]{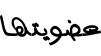}}
    \end{minipage}
    \hspace{0.01\linewidth}
    \begin{minipage}[b]{0.2\linewidth}
        \centering
        \frame{\includegraphics[height=0.5cm]{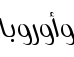}}
    \end{minipage}
    \hspace{0.01\linewidth}
    \begin{minipage}[b]{0.2\linewidth}
        \centering
        \frame{\includegraphics[height=0.5cm]{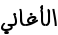}}
    \end{minipage}
    \caption{AMFDS Dataset [\ref{28}] raw data of words}
    \label{fig: raw data}
    \vspace{10pt}
    \centering
    \includegraphics[width=1\linewidth]{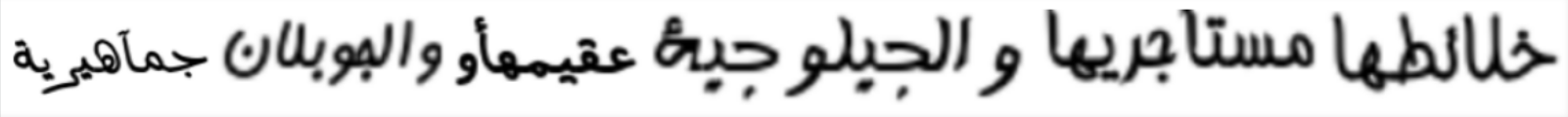}
    \caption{Generated line from the raw words data }
    \label{fig: Sample generated line}
    \vspace{-15pt}
\end{figure}

We used the Arabic Multi-Fonts Dataset (AMFDS)[\ref{28}] for our OCR problem. Our sample raw data of the Dataset has images of one word with 18 different fonts with 2 million words. Word size varies between 7 to 10 characters in the raw data as shown in Figure \ref{fig: raw data}. We Concatenate words horizontally at random to form a sentence for training the OCR model all done dynamically using a generator to leave no room for duplicate sentences. Figure \ref{fig: Sample generated line} shows a sample of the generated line. The generator has an augmentation capability applying Dilation, Rotation, and Salt effect as in Figure \ref{Salted sample}, and Boldness effect in Figure \ref{Bolded sample}. The Salt effect is done for comparison purposes with the related work tested on the same dataset. As for the Boldness filter, it simulates hard strokes in the real-world images.

\subsubsection{Training}
Our training data generator generates sentences that contain a random number of words with a specified max number. We train the model on 30:70 Augmented to Clean Data ratio. The Augmentation processes are initialized randomly on each line. Throughout our experiments, we saw that it's not possible to train the OCR model on a long sequence of words right from the beginning. Therefore, We begin the training with a small (1 or 2 words max) then by applying Transfer Learning techniques we freeze the initial CNN Layers "responsible for Feature Extraction" and we train the last BLSTM with the Softmax on the increased sequence of words. We increased the maximum number of words by 2 each time until reached an optimal solution. 
\section{Results And Discussion}
\subsection{Line Segmentation}
The difference between the initial weights "best-weights of the DBNet++ [\ref{11}] Model" and our Fine-Tuned weights on The Arabic dataset can be seen in Table \ref{table1}. The two models were tested on Customized Arabic Handwritten and Non-Handwritten data with precise Annotations on the line level.
The evaluation of the model is using the following metrics: 
\begin{itemize}
    \item {Precision: measure how accurately the model segments the lines.}
    \item {Recall: measure how effectively the model captured most of the lines in the images.}
    \item {-F-Measure: balance performance in terms of both accuracy and completeness of the line segmentation.}
\end{itemize}

\begin{table*}[h!]
\centering
\caption{OCR Model Performance Results}
\label{OCR Module Performance Results table}
\vspace{-7pt}
\begin{tblr}{
  colspec = {|c|c|cc|cc|cc|cc|cc|c|},
  hlines,
  vlines,
  colsep = 2pt,
  rowsep = 0.8pt
}
\# & Num of Words & \SetCell[c=2]{c}{Solid Accuracy\%} & & \SetCell[c=2]{c}{Salted Accuracy\%} & & \SetCell[c=2]{c}{Bolded Accuracy\%} & & \SetCell[c=2]{c}{Solid Fasha[\ref{28}]\%} & & \SetCell[c=2]{c}{Salted Fasha[\ref{28}]\%} & & Notes \\ 
   & & CRR & WRR & CRR & WRR & CRR & WRR & CRR & WRR & CRR & WRR & \\ 
1  & 1 & 99.20 & 93.75 & 85.26 & 31.87 & 88.85 & 37.68 & 98.81 & 90.53 & 82.01 & 21.48 & (18) fonts, unique words dataset \\ 
2 & 2 & 93.67 & 62.18 & 87.11 & 34.84 & 86.75 & 33.85 & - & - & - & - & - \\ 
3 & 3 & 93.70 & 60.20 & 88.23 & 37.39 & 87.79 & 35.59 & - & - & - & - & - \\ 
4 & 4 & 84.15 & 31.42 & 80.41 & 24.61 & 80.25 & 24.61 & - & - & - & - & - \\ 
5 & 5 & 81.29 & 30.87 & 71.88 & 17.225 & 70.25 & 16.62 & - & - & - & - & - \\ 
6 & 6 & 66.01 & 18.78 & 64.95 & 13.48 & 63.50 & 14.39 & - & - & - & - & - \\ 
\end{tblr}
\vspace{-10pt}
\end{table*}

\begin{table}[htbp]
\caption{Improvment Detecting Arabic Handwritten and printed lines}
\vspace{-12pt}
\begin{center}
\begin{tabular}{|c|c|c|c|}
\hline
\textbf{Model} & \textbf{Precision} & \textbf{Recall} & \textbf{F-Measure} \\
\hline
Universal Model & 61.53 & 34.60 & 41.33 \\
\hline
Our Model & 81.66 & 78.82 & 79.07 \\
\hline

\end{tabular}
\label{table1}
\end{center}
\end{table}

\begin{figure}[htbp]
    \centering
    \vspace{-15pt}
    \begin{minipage}[b]{0.3\linewidth}
        \centering
        \includegraphics[width=\linewidth]{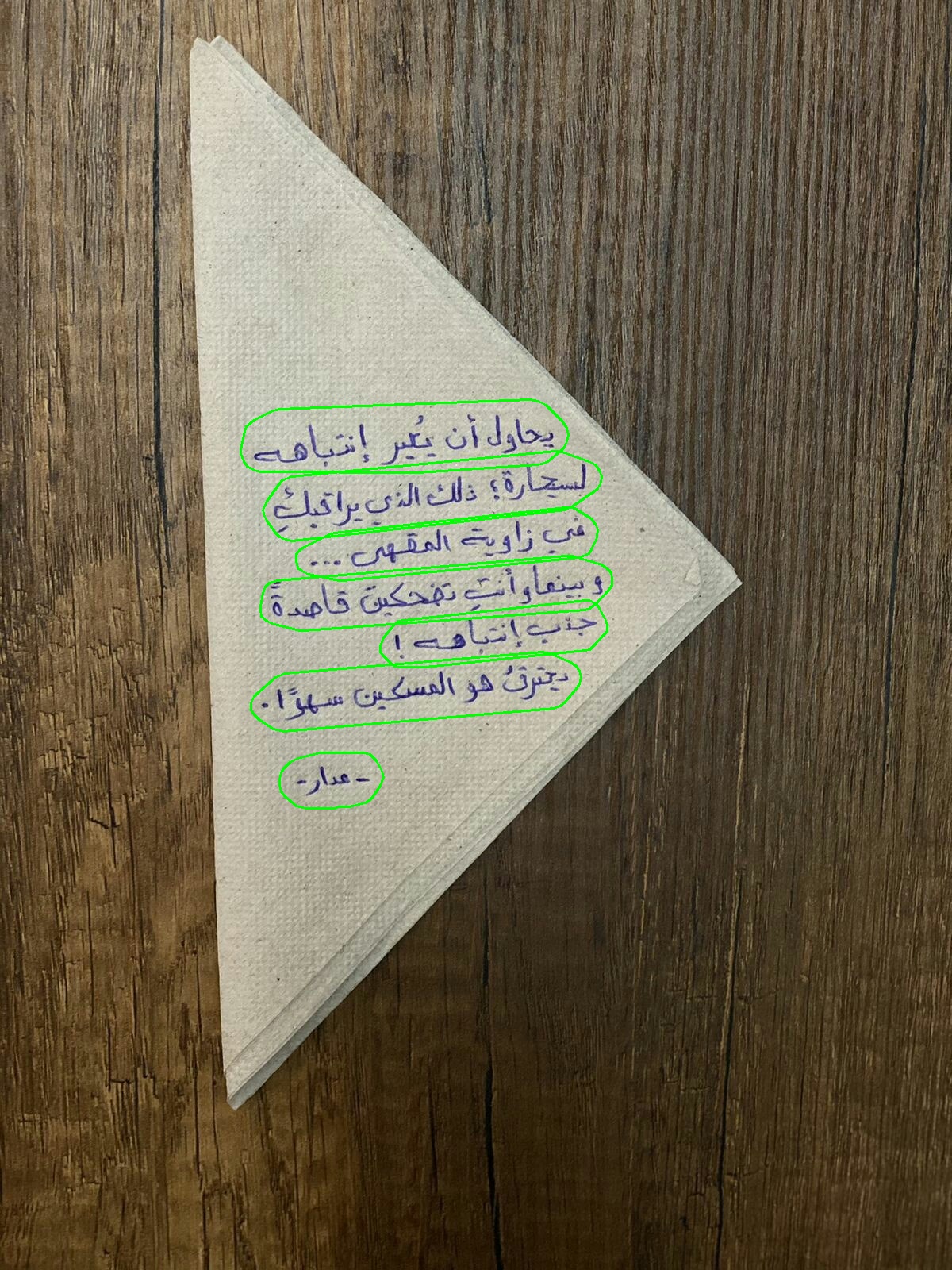}
        \label{fig:1-after}
    \end{minipage}
    \hspace{0.01\linewidth}
    \begin{minipage}[b]{0.41\linewidth}
        \centering
        \includegraphics[width=\linewidth]{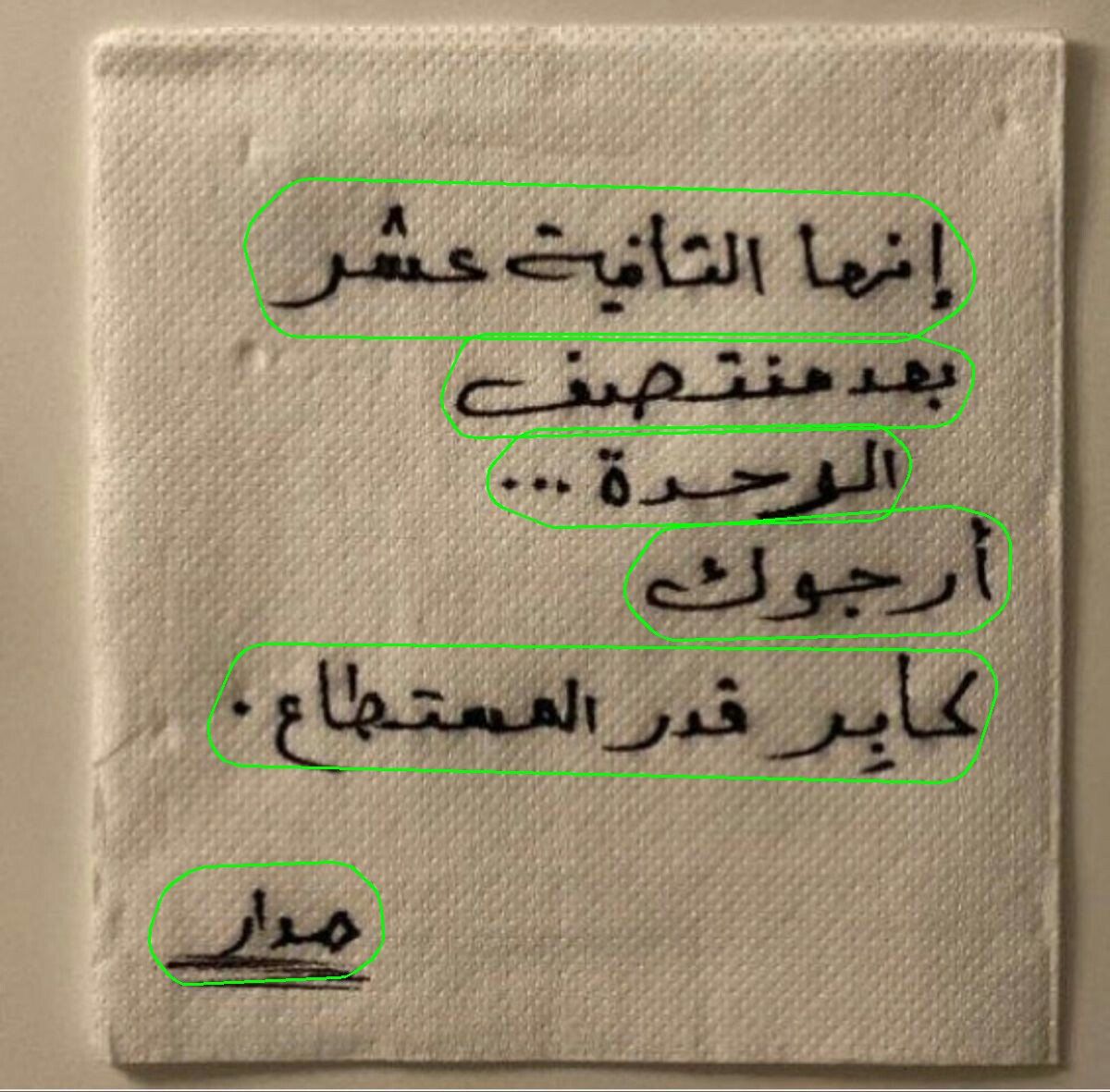}
        \label{fig:3-after}
    \end{minipage}
    \vspace{-15pt}
    \caption{Visualizing Line Segmentation Results}
    \vspace{-10pt}
\end{figure}

\subsection{OCR Engine}
\begin{figure}[htbp]
    \centering
    \begin{minipage}[b]{0.15\textwidth}
        \centering
        \includegraphics[width=\textwidth]{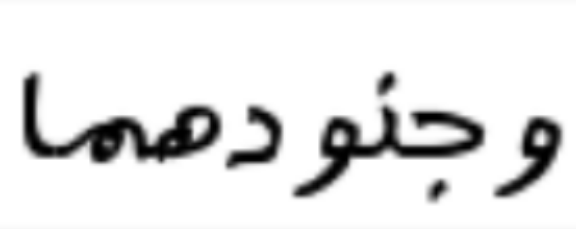}
        \caption{Solid sample}
        \label{Solid sample}
    \end{minipage}
    \hfill
    \begin{minipage}[b]{0.15\textwidth}
        \centering
        \includegraphics[width=\textwidth]{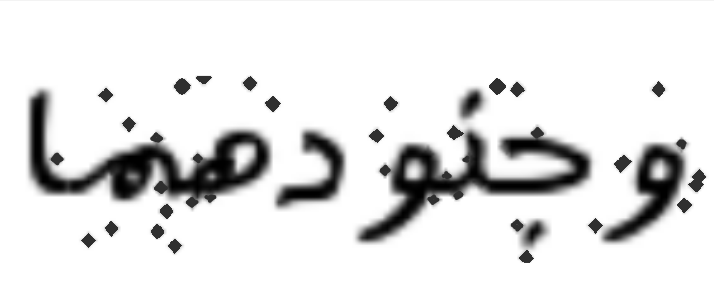}
        \caption{Salted sample}
        \label{Salted sample}
    \end{minipage}
    \hfill
    \begin{minipage}[b]{0.15\textwidth}
        \centering
        \includegraphics[width=\textwidth]{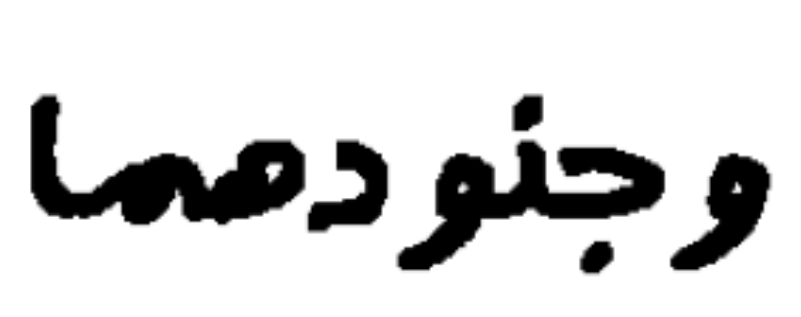}
        \caption{Bolded sample}
        \label{Bolded sample}
    \end{minipage}
    
    \vspace{-12pt}
\end{figure}

We tested our models on 3 modes Solid in Figure \ref{Solid sample}, Salted in Figure \ref{Salted sample}, and Bold in Figure \ref{Bolded sample} with different words count for all sentences in the experiments as shown in Table \ref{OCR Module Performance Results table}. The results obtained on the 3 modes are very similar for the same number of words. As we increase the sequence of words in the sentence, we noticed that the model struggles to recognize the middle words of the sentence, unlike the edged words due to the weak representation of words in the Recurrent Layer. WRR results are decreasing rapidly because the dataset words contain 7 to 10 characters which means hard tesing dataset. As one character miss would result in a word miss-recognized, therefor low WRR score. As for comparison, our results overcomes Fasha et al[\ref{28}] results on the AMFDS dataset on one-word as well as Salted filter as shown in table\ref{OCR Module Performance Results table}.
\vspace{-8pt}

\section{Conclusion}
Arabic Handwritten Text Recognition remains a major challenge because of the cursive nature of Arabic script, the diversity of handwriting styles, and the unavailability of large labeled data sets.
This paper addressed these issues by presenting a comprehensive solution that enhances line segmentation accuracy. It presents a novel combination of differentiable binarization and adaptive scale fusion techniques, Integrated with a CNN-BiLSTM-CTC OCR Model. Our detection model has generally proven useful in segmenting Arabic script lines correctly. However, the efficiency of the OCR engine degrades when dealing with longer text sequences, indicating a need for further optimization in handling extended text lengths. \\ Future work would be to achieve an optimization of the model on longer text sequences without much loss in accuracy by developing a representation of handwritten words more independent of sequence length.

\end{document}